\pdfoutput=1

\documentclass[11pt]{article}

\usepackage[table]{xcolor}

\usepackage[final]{acl}

\usepackage{times}
\usepackage{latexsym}
\usepackage{listings}
\usepackage{multicol}
\usepackage[T1]{fontenc}
\usepackage{multirow}
\usepackage{array}
\usepackage{float}
\usepackage{booktabs} 
\usepackage{csquotes}
\usepackage[utf8]{inputenc}
\usepackage{enumitem}
\usepackage{microtype}
\usepackage{hyperref}
\usepackage{tabularx}

\usepackage{inconsolata}
\usepackage{dsfont}
\usepackage{graphicx}
\usepackage{algorithm}
\usepackage{algorithmic}
\usepackage{adjustbox}
\usepackage{amsmath}
\usepackage{pifont}
\usepackage[acronym]{glossaries}
\newcommand{\cmark}{\ding{51}}%
\newcommand{\xmark}{\ding{55}}%
\newacronym{uoc}{UOC}{Unified Opinion Concept}
\newacronym{uoce}{UOCE}{Unified Opinion Concept Extraction}
\newacronym{acos}{ACOS}{Aspect Category Opinion Sentiment}
\newacronym{absa}{ABSA}{Aspect-Based Sentiment Analysis}
\newacronym{aste}{ASTE}{Aspect Sentiment Triple Extraction}
\newcolumntype{?}{!{\vrule width 1pt}}

\definecolor{prop_col}{RGB}{215, 250, 245} 
\definecolor{class_col}{RGB}{109, 177, 255} 
\title{Towards Semantic Integration of Opinions: Unified Opinion Concepts Ontology and Extraction Task}

\author{Gaurav Negi, Dhairya Dalal, Omnia Zayed, \and Paul Buitelaar \\
    Insight SFI Research Centre for Data Analytics\\
        Data Science Institute\\
        University of Galway \\ 
         \{gaurav.negi, omnia.zayed, paul.buitelaar\}@insight-centre.org,\\ d.dalal1@universityofgalway.ie}

\begin{document}

\maketitle
\begin{abstract}
This paper introduces the Unified Opinion Concepts (UOC) ontology to integrate opinions within their semantic context. The UOC ontology bridges the gap between the semantic representation of opinion across different formulations. It is a unified conceptualisation based on the facets of opinions studied extensively in NLP and semantic structures described through symbolic descriptions. 
We further propose the Unified Opinion Concept Extraction (UOCE) task of extracting opinions from the text with enhanced expressivity. 
Additionally, we provide a manually extended and re-annotated evaluation dataset for this task and tailored evaluation metrics to assess the adherence of extracted opinions to UOC semantics. Finally, we establish baseline performance for the UOCE task using state-of-the-art generative models. 
\end{abstract}

\section{Introduction}
Opinion\footnote{We use the term opinion
as a broad concept that covers sentiment and its
associated information such as opinion target and the person who holds the opinion,
and use the term sentiment to mean only the underlying positive, negative or neutral polarity
implied by opinion.} mining has seen a move from a traditional sentence- and document-level analysis \cite{DBLP:conf/emnlp/PangLV02} to fine-grained approaches. 
Aspect-based Sentiment Analysis (ABSA) is a notable approach for fine-grained opinion mining, and it has been extensively studied in natural language processing (NLP) research. \cite{DBLP:conf/semeval/PontikiGPPAM14,pontiki-etal-2015-semeval,DBLP:conf/semeval/PontikiGPAMAAZQ16,fiqa_task}.  The task focuses on identifying the aspects of the entities and their associated sentiments from a given text sequence. In the following sentence:

\begin{displayquote}
\vspace{-8pt}
\textit{``I had hoped for better \underline{battery life} , as it had only about 2-1/2 hours doing heavy computations (8 threads using 100 \% of the CPU) ."}
\end{displayquote}  
\noindent ABSA results are extracted as the following tuple: \{\underline{battery life}, Battery\#Operational\_Performance, negative\}. The extracted tuple is in the form \{aspect term/opinion target, entity\#aspect category, sentiment polarity\}. Opinion target (often called aspect term) is the word or phrase over which an opinion is expressed. The aspect category is an attribute of the opinion target, and sentiment polarity specifies whether the opinion is positive, negative, or neutral. This fine-grained analysis allows for a more detailed understanding of opinions and sentiments expressed in the text.

\begin{figure}[ht]
    \centering
    \vspace{-10pt}
    \includegraphics[width=0.45\textwidth, height=1.75cm]{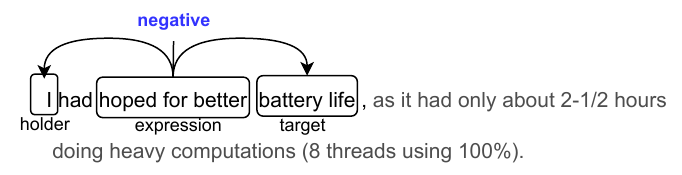}
    \caption{Structured Sentiment Analysis }
    \vspace{-6pt}
    \label{fig:struct_senti}
\end{figure}

Structured sentiment analysis \cite{structured_sentiment_barns_2022} is another formulation of opinion mining, where the nodes are spans of sentiment holders, targets and expressions, and the arcs are the relations between them. Figure \ref{fig:struct_senti} illustrates this formulation.

\noindent ABSA and structured sentiment analysis overlap significantly in extracting specific opinion facets\footnote{We use the term facet as used by Bing Liu to describe various subtasks and the building blocks of an opinion.}. None of the formulations fully incorporate all opinion facets proposed by \citet{liu2017many}, which reduces the expressiveness and granularity of the extracted opinions. The example above shows that the opinion is valid for specific individuals or groups engaged in " \textit{doing heavy computations}". The reason for opinion is also expressed, i.e. \textit{``it had only about 2-1/2 hours"}. None of the existing opinion mining formulations enable these extractions. 

This work investigates semantic representations of opinions to enrich their expressiveness. Towards this end, we studied the specification of opinion for the Semantic Web as described by the Marl Ontology \cite{marl_opinion_2011}. However, it has a limited cross-compatibility with the opinion formulations researched in NLP. We unify the opinion facets studied extensively in NLP with the semantic structures described in the Marl Ontology to develop a comprehensive Unified Opinion Concepts (UOC) framework. The UOC ontology consolidates and formalises these opinion components into an exhaustive set, enabling the semantic representation of opinions in a structured and unambiguous manner. UOC leverages the implicit hierarchies and relationships across diverse NLP frameworks based on the theoretical foundations of \citet{DBLP:books/sp/mining2012/LiuZ12}.
\noindent Our contribution\footnote{Github Repository: \url{https://github.com/gauneg/UnifiedOpinionConcepts_LDK_2025}} can be summarized as follows:

\begin{itemize}
\vspace{-4pt}
  
    \item We introduce the UOC ontology (Section \ref{sec:semantic_ontology}) that conceptualises semantic representation of an opinion, improving on the existing opinion formulations in terms of expressivity and cross-compatibility. 
    \vspace{-6pt}
    
    \item We define Unified Opinion Concept Extraction (UOCE) as an NLP task (Section \ref{sec: prob_def}) grounded in the rich semantic representation of the UOC ontology (Section \ref{sec:bline}).    
    \vspace{-6pt}
    
    \item We extend annotations of an existing gold standard opinion mining dataset (Section \ref{sec: dataset}) to create an evaluation dataset for UOCE. We propose tailored evaluation metrics (Section \ref{sec:eval_metrics}) for rigorous baseline assessment.

\end{itemize}

\section{Related Work}
\noindent \textbf{Opinion Mining in NLP.}
ABSA evolved from feature-based summarisation \cite{hu_mining_2004,zhuang_movie_2006,ding2008liu} and the foundational work on opinion mining by \citet{DBLP:books/sp/mining2012/LiuZ12}, which involves extracting and summarising opinions on features (attributes/keywords). The downstream tasks that spun out of the ABSA research space can be classified into the following categories based on the opinion facets they address: Opinion Aspect Co-extraction \cite{qiu_opinion_2011,liu_opinion_coextratcion_2013,liaspect,wang_coupled_2017}, Aspect Sentiment Triple Extraction (ASTE) \cite{aste_ote_mtl,aste_jet,grid_at_op_sentiment_2020}, Target-Aspect-Sentiment Detection (TASD) \cite{tasd_tas_bert, tasd_mejd}, Aspect-Category-Opinion-Sentiment (ACOS/ASQP) quadruple extraction \cite{acos_extract_classify, mvp_2023,acos_gen_nat_scl_bart}. \citet{barnes_structured_2021, barnes_flaunt_senti_2021} perform opinion tuple extraction as dependency graph parsing, where the nodes are spans of sentiment holders, targets and expressions, and the arcs are the relations between them (see Figure. \ref{fig:struct_senti}). 
We extend these existing opinion formulations by adding more elements to increase expressivity and formalise the relationships between opinion facets with an ontology.
 


\paragraph{Ontological Methods.}

Ontologies provide an explicit machine-readable specification of shared conceptualization, and our inquiry into existing ontologies for opinion expression led us to the Marl Ontology\footnote{https://www.gsi.upm.es/ontologies/marl/}\cite{marl_opinion_2011}. It is a standardised schema designed to annotate and describe subjective opinions expressed on the Semantic Web and in information systems \cite{senpy_2016,linguistic_linked_data_2013}. However, the Marl ontology cannot describe fine-grained opinion mining currently being researched in NLP. \citet{kim_onto_2018} propose a task ontology to facilitate sentiment classification of the given aspect terms; it does not contribute towards highlighting fine-grained opinion representation. Our work reformulates and extends the domain ontology of an opinion, improving the interfacing of opinion description across different disciplines. 
\paragraph{Neuro-Symbolic Methods.} Sentiment Analysis with neuro-symbolic methods adds knowledge and symbolic constraints to assist the deep learning models. This knowledge can be in the form of structured linguistic characteristics with WordNet, SentiWordNet \cite{multidl_neuro_sym_2022}, word-sense disambiguation \cite{wsd_multi_2022, wsd_neuro_sym_2023} or using domain-specific knowledge \cite{domain_neuro_sym_2023}. Neuro-symbolic work on opinion mining does not extend or introduce novel formulations of opinion-mining tasks. 

We utilise the nuances of opinion-mining literature to reformulate the opinion ontology to bridge the gap between the differences in the semantic conceptualisation of opinion expression. We align the concepts of Marl with the various opinion mining NLP tasks (i.e. ASTE, TASD, ACOS, structured sentiment analysis) and the implicit hierarchies within these conceptualisation frameworks. We propose the ontology, a benchmark dataset, and the baseline methods for opinion extraction.

\begin{figure*}[btp]
    \centering
    \includegraphics[width=\textwidth]{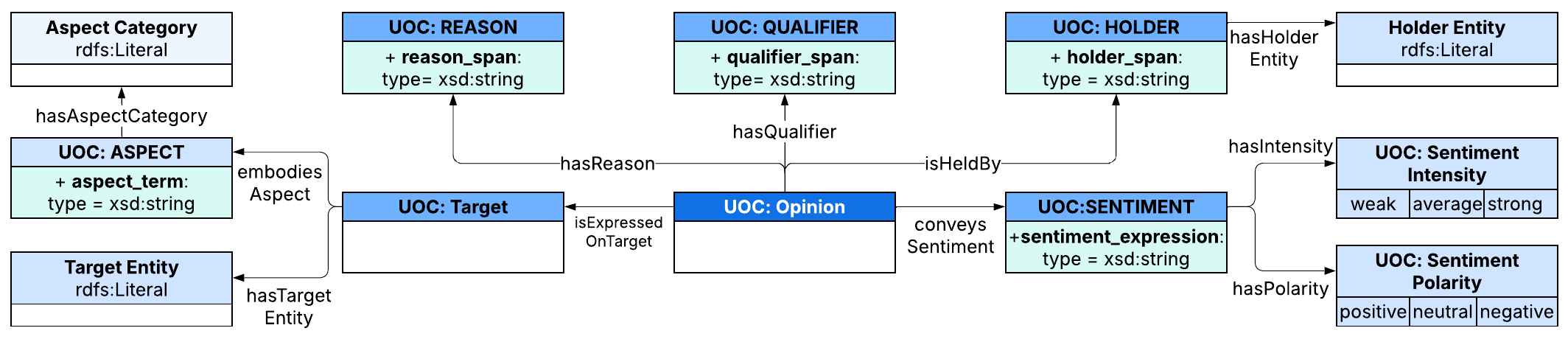}
    
    \caption{Unified Opinions Concepts (UOC) Ontology Diagram}
    \label{fig:Uoc}
    \vspace{-15pt}
\end{figure*}

\section{Unified Opinion Concept Ontology} 
\label{sec:semantic_ontology}
One of the primary objectives of this work is the development of an ontology to describe opinions and the associated semantics precisely. An ontology is an explicit, machine-readable specification of a shared conceptualisation. The UOC ontology shown in Fig.\ref{fig:Uoc} describes the following components: (i) \colorbox{class_col}{Classes} conceptualising opinion and its facets, (ii) \colorbox{prop_col}{Attributes} of classes along with the datatype property (+attribute\_name:type= datatype property), and (iii) object properties, that describe relationships between the concepts represented by the classes. 

We formalize the ontology of opinions through a two-step process. First, we identify tasks within the domain of opinion mining and examine the overlap of their facets with the concepts in the Marl ontology. These facets and concepts are then
 aligned and integrated to establish a unified representation of opinion concepts, we refer to as Unified Opinion Concepts (UOCs).  Table \ref{tab: concept_data_align} shows the concept alignments and the resulting UOCs.
 
 \begin{table}[!ht]
\setlength\extrarowheight{1pt}
\addtolength{\tabcolsep}{.1pt}
\scriptsize
    \begin{tabular}{p{0.22\linewidth}|p{0.3\linewidth}|p{0.3\linewidth}}
     \hline
     \textbf{Marl Ontology} & \textbf{NLP Frameworks} &\textbf{UOC} \\
     
      \hline
        Polarity Value& Sentiment Intensity & Sentiment Intensity\\
        \hline
         Polarity Class & Sentiment Orientation & Sentiment Polarity \\
         \hline
         Opinion Text & Sentiment Expression & Sentiment Expression  \\
         \hline
         Described  & Aspect Category & Aspect Category\\
         Object Feature & &    \\
         \hline
         Described  & Opinion Target / & Aspect Term \\
         Object Part & Aspect term & \\
         \hline
         Described Object & Entity & Target Entity \\
         \hline
         NA & Opinion Time (t) & NA  \\
         \hline
         NA & Opinion Qualifier & Qualifier \\
         \hline
         NA & Opinion Reason & Reason \\
         \hline
         \multirow{2}{*}{NA} & \multirow{2}{*}{Opinion holder} & Holder Entity \\ \cline{3-3}
            & & Holder Span \\ \hline
    \end{tabular}
    \caption{Unified Opinion Concepts (UOC)}
    \label{tab: concept_data_align}
\end{table}
\vspace{-8pt}
 Second, we leverage the explicit and implicit hierarchical structures described in the NLP literature to define the relationships between these concepts, thereby formalizing the UOC ontology. 
 

\subsection{Modelling Ontological Concepts and Relationships}

\noindent We examine the conceptualization of opinion facets and explore how insights from NLP research shapes the ontology development process. \citet{liu2017many}, posits that an opinion comprises two fundamental components: sentiment and target. This conceptualization is reflected in the proposed ontology as shown in Figure \ref{fig:Uoc}. The individual concepts introduced in Table \ref{tab: concept_data_align} and their associated properties are discussed below.
 
\begin{figure}[ht]
     \vspace{-8pt}
    \centering
    \includegraphics[width=0.9\linewidth, height=1.95cm]{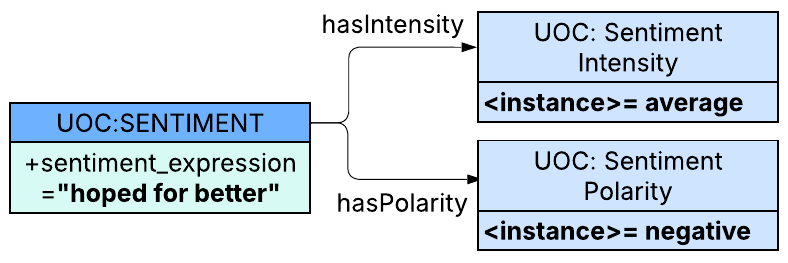}
    \vspace{-5pt}
    \caption{\textbf{UOC Sentiment} extracted from: \textit{"I had \underline{hoped for better} battery life , as it had only about 2-1/2 hours doing
heavy computations (8 threads using 100 \% of the CPU)"}}
    \label{fig:uoc_sentiment}
    \vspace{-10pt}
\end{figure}

\smallskip \noindent \textbf{Sentiment}: This class encapsulates the underlying feelings expressed in an opinion. It is composed of several interconnected concepts that collectively define Sentiment. The relationship between Sentiment and Opinion is articulated using the object property \textit{conveysSentiment}. The semantic structure of the Sentiment class reflects its strong agreement with structured sentiment analysis formulation. Figure \ref{fig:uoc_sentiment} illustrates an instance of the Sentiment class, its constituents, and their relationships. Its key components—Sentiment Intensity, Sentiment Polarity, and Sentiment Expression—are defined as follows:

    \begin{enumerate}
        \item \textbf{Sentiment Intensity}: This component captures the strength of the identified sentiment expressed in an opinion.  For this study, we represent intensity using discrete ordinal values: weak$<$average$<$strong. It corresponds to the Polarity class of the Marl ontology and sentiment intensity of the NLP opinion frameworks. The relationship between \textbf{Sentiment Intensity} and \textbf{Sentiment} is defined by the property \textit{hasIntensity}.
        
        \item \textbf{Sentiment Polarity}: This refers to the predefined semantic orientation of a sentiment (i.e. positive, negative or neutral). Marl also uses the class Polarity in the ontology to represent the concept. In contrast, NLP frameworks sometimes identify it as sentiment orientation. The \textit{hasPolarity} property associates \textbf{Sentiment} with this component.
        \item \textbf{Sentiment Expression}: The Sentiment Expression is the subjective statement that indicates the presence of a sentiment, often explicitly appearing as a word or phrase in the text. In ABSA, this facet is frequently referred to as "opinion", "opinion text", or "opinion span". However, as structured sentiment analysis posits, sentiment expression is more strongly associated with sentiments, particularly in this more fine-grained form of analysis with further disambiguation between the sentiment and the target of an opinion. In the UOC ontology, it is an attribute of the \textbf{Sentiment} class.
    \end{enumerate}

    \noindent \textbf{Target}: This class encapsulates the subjective information on which an opinion is expressed. It represents a composite concept comprising fine-grained components that collectively define the Aspect and Entity implicated in the opinion.
    This conceptualization is in agreement with the ABSA literature. Figure \ref{fig:uoc_target} illustrates an instance of the Target class, its constituents, and their relationships. It addresses the semantic formulation for extracting the multiple facets of an opinion's target. The object property \textit{isExpressedOnTarget} describes its relationship with \textbf{Opinion} class. The conceptualization of \textbf{Target}is described as follows:
    \begin{enumerate}
    
    \item \textbf{Target Entity}: It is the object of interest on which a sentiment is explicitly or implicitly expressed. It may refer to a product, service, topic, issue, person, organization, or event. While traditional ABSA datasets often conflate entities with aspect categories, we define \textbf{Target Entity} as an independent concept, motivated by advancements in Entity-Level Sentiment Analysis \cite{elsa_2022}, which broadens its scope and applicability. The relationship between \textbf{Target} and \textbf{Target Entity} is represented by the property \textit{hasTargetEntity}. The \textbf{Target Entity} can take two forms: as an "xsd: string" or an Internationalized Resource Identifier\footnote{IRIs are particularly useful for connecting concepts to a knowledge graph on the Semantic Web.} (IRI).

    \item \textbf{Aspect}: Aspect describes the part and attribute of \textbf{Target Entity} on which the sentiment is expressed. The \textit{embodiesAspect} property describes its relationship to \textbf{target}. It is semantically deconstructed into the following sub-units:
    \begin{enumerate}
        \item \textbf{Aspect Category}: It expresses attributes or properties of the aforementioned \textbf{target}. Its relationship to Aspect is described by \textit{hasAspectCategory}. This class can be instantiated in two forms. The category can be described as an "xsd:string" data property or as an IRI.
        \item \textbf{Aspect Term}: An explicit expression (e.g., words or phrases) in the input text indicates an aspect category. It is an attribute of the \textbf{Aspect} class.
    \end{enumerate}

        \begin{figure}[ht]
    \centering
    \includegraphics[width=0.9\linewidth]{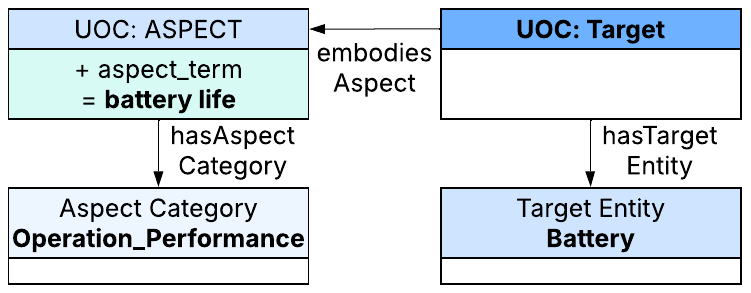}
    \vspace{-8pt}
    \caption{\textbf{UOC Target} extracted from: \textit{"I had hoped for better \underline{battery life} , as it had only about 2-1/2 hours doing
heavy computations (8 threads using 100 \% of the CPU)"}}
    \label{fig:uoc_target}
    \vspace{-10pt}
\end{figure}

    \end{enumerate}
    
     \noindent \textbf{Holder}: An opinion holder (an opinion source) is a person or organization expressing an opinion. The relationship of the \textbf{Holder} class with \textbf{Opinion} is described by \textit{isHeldBy} property. A counterpart for the opinion holder in Marl ontology does not exist. It is expressed in the UOC ontology by the use of the following hierarchical sub-components:
    \begin{enumerate}
        \item \textbf{Holder Entity}: It corresponds to the individual or organization articulating the opinion. These entities may include persons, organizations, products, or other entities relevant to the opinion context. The \textit{hasHolderEntity} property describes its relationship with the \textbf{Holder} class.
        \item \textbf{Holder Span}: It is an attribute of the \textbf{Holder} class and comprises the actual words or phrases in the text indicating the \textbf{Holder} of an \textbf{Opinion}.
    \end{enumerate}

    \noindent \textbf{Qualifier}: A Qualifier refines the scope or applicability of an opinion, delineating the group or subgroup to which the opinion pertains. For instance, in the sentence:
    
    \textit{``I had hoped for better battery life, as it had only about 2-1/2 hours \colorbox{prop_col}{doing heavy computations} (8 threads using 100 \% of the CPU)"} 
    
    The qualifier \textit{``doing heavy computations"} specifies the subset for whom the battery life would be inadequate. The property \textit{hasQualifier} describes the relationship between \textbf{Opinion} and \textbf{Qualifier}.
    
    \medskip \noindent \textbf{Reason}: A reason represents an opinion's justification or underlying cause. This concept is connected to the \textbf{Opinion} class via the property \textit{hasReason} and, like \textbf{Qualifier}, only existed as a theoretical construct in NLP research.
    
    e.g. \textit{``I had hoped for better battery life, as \colorbox{prop_col}{it had only about 2-1/2 hours} doing heavy computations (8 threads using 100 \% of the CPU)"}
    
    It has the reason for the opinion which specifically addresses the battery issues, i.e. \textit{``it had only about 2-1/2 hours"}
    
    Only the explicit reasons stated within the text are considered for this study. Implied reasons, although they may exist, are not taken into account for this work.

\section{Unified Opinion Concept Extraction (UOCE)}
We harness the rich semantics of the UOC ontology to propose Unified Opinion Concept Extraction (UOCE), an NLP task for comprehensive opinion extraction. To facilitate UOCE solutions, we provide (i) the formalized problem definition, (ii) the evaluation metrics, (iii) the analysis of existing datasets and the extension of annotations for method evaluation, and (iv) baseline methods with LLMs.
\subsection{Problem Definition}
\label{sec: prob_def}
Given an input text $T_i$, extract an exhaustive set of opinions $O_i = \{o_{i,j}|j=1,2...|Oi| \}$ where each opinion $o_{i,j}$ is represented as  tuple:

\vspace{-10pt}
\begin{equation}
   \label{eq: form_op} 
   \begin{aligned}
    o_{i,j} = (at_{i,j}, ac_{i,j}, te_{i,j}, se_{i,j}, sp_{i,j}, \\
    si_{i,j}, hs_{i,j}, he_{i,j}, q_{i,j}, r_{i,j})\\
    \text{or using the shorthand notation as follows:}\\
    o_{i,j} = (at, ac, te, se, sp, si, hs, he, q, r)_{i,j}
    \end{aligned}
\end{equation}

\begin{table}[ht]

\vspace{-10pt}
\small
\begin{tabular}{l l }    
     \textbf{where:} &  \\
      $at$: aspect term, & $ac$: aspect category, \\
     $te$: target entity,& $se$: sentiment expression, \\ 
    $sp$: sentiment polarity, & $si$: sentiment intensity, \\
    $hs$: holder span, & $he$: holder entity, \\
    $q$: qualifier, & $r$: reason
\end{tabular}
\end{table}

\noindent Each tuple encapsulates the key components necessary to define an opinion. This NLP task is formulated on the UOC semantics described in Section \ref{sec:semantic_ontology}, making it possible to instantiate knowledge graphs from the extract opinion(s) using the UOC schema.

\subsection{Evaluation Metrics}
\label{sec:eval_metrics}
The selection of the evaluation metrics is informed by the ability to measure the following: (i) The agreement with the ground truth across the extracted opinion tuples, (ii) The agreement with the ground truth of individual elements of extracted opinions, (iii) Metrics used by state-of-the-art opinion mining systems for fair comparison.  

\paragraph{Tuple-Level Exact Match Metric}
A predicted tuple of opinion components is considered correct only if all the individually extracted components exactly match the ground truth. Precision, recall and F1 scores are calculated with this intuition for the exact match of all the elements in the tuple. The tuple-level exact match metrics evaluate many fine-grained opinion mining systems.
.\cite{grid_at_op_sentiment_2020, acos_extract_classify, aste_jet, tasd_aste_gas}.

\paragraph{Component-Level Exact Match Metric}
The tuple-level exact match metric severely penalizes the mismatch in the measured values; even a slight mismatch of one component completely devalues the entire extracted opinion. In doing so, it does not account for the partially correct extracted opinions, exacerbating the non-linearity or discontinuity of the evaluation metrics discussed in elaborate detail by \citet{nemerge_2023}. Therefore, our metric of choice is the Component-level exact match metric discussed in the remainder of this section.

In the dataset with text instances $\{T_i\}_{i=1}^N$ for each text instance $T_i$ there exists the ground truth opinion annotation $Og_i$ is a set of opinions $Og_i = \{og_{i,j}|j=1,2,...,|Og_i|\}$ and the corresponding set of predicted opinions $Oe_i=\{oe_{i,k}|k=1,2,...|Oe_i|\}$. Each opinion instance has ten components described in eq. \ref{eq: form_op}. For any pair of tuples $(oe_{i,k}, og_{i,j})$ we describe the degree of agreement as:

$$f(oe_{i,k}, og_{i,j}) = \frac{|oe_{i,k} \cap og_{i,k}|}{|og_{i,k}|}$$
We perform a one-to-one matching (without replacement) between the tuples in $Oe_i$ and $G_i$. Now $\mathcal{A}_i \subseteq Og_{i} \times Oe_{i}$, is the set of aligned tuple pairs obtained. For each gold tuple $og_i\in G_i$ at most one predicted/extracted tuple is selected (without replacement, one predicted tuple cannot be matched with other ground truth tuples.). The selection can also be shown as:
$$\mathcal{A}_i = \arg \max_{\mathcal{M}\subseteq Og_i \times Oe_i \text{matching}} \sum_{og,oe \in \mathcal{M}} f(oe, og)$$
Any extracted tuple not included in $\mathcal{A}_i$ does not contribute towards true positive. However, it does bring precision down as it is considered when counting the total extracted opinion tuples. Now for each text input $T_i$ we calculate true positive $$TP= \sum_{i=1}^{N} \sum\limits_{(og,oe)\in \mathcal{A}_i}f(oe, og)$$. Precision $P$ and recall $R$ are then given by:
$$P=\frac{TP}{\sum_{i=1}^{N} |Oe_i|} \text{ , } R= \frac{TP}{\sum_{i=1}^{N}|Og_i|}$$

The combined metrics account for the presence/absence of the extracted opinion(s) in the annotated opinion(s) and the degree of agreement between the extracted opinion components and the ground truth. The two metrics are compared in the Appendix \ref{sec:appendix_metric}. 


\subsection{Dataset}
\label{sec: dataset}
We use the semantic structure of opinion defined by the UOC ontology to create an evaluation dataset. The dataset includes annotations for components listed in Eq \ref{eq: form_op}. We annotate the evaluation dataset in two steps: (i) Semantic validation of the labels of the existing dataset based on UOC Ontology. (ii) Using the outcome of the semantic validation to select and extend the annotations.

\subsubsection{Semantic Data Validation}
\label{sec:bench_data}
The mappings in Table \ref{tab:sem_data_alignment} highlight the opinion mining datasets and the corresponding annotations for the opinion facets. We evaluate the suitability of a dataset for the UOCE task through this semantic assessment. The datasets in the table are listed across the top row, while different concepts are listed in the first column. A check mark \ding{51} indicates a dataset's agreement with the UOC ontology for a specific concept.

\begin{table}[ht]
\footnotesize
\renewcommand{\arraystretch}{1.15} 
\resizebox{0.485\textwidth}{!}{
\begin{tabular}{l|c|c|c|c|c|c|c}
\hline
\textbf{Datasets} & $si$ & $sp$ & $se$  & $ac$  & $at$  & $te$  & $hs$ \\
\hline
$D_{10}$\cite{DBLP:conf/acl/ToprakJG10}&  & \cmark & \cmark &  & \cmark &   & \cmark \\
\hline
$SL_{14}$ \cite{DBLP:conf/semeval/PontikiGPPAM14} &  & \cmark &  & \cmark & \cmark &    &  \\
\hline
$SR_{14}$ \cite{DBLP:conf/semeval/PontikiGPPAM14}&  & \cmark &  &  & \cmark &    &  \\
\hline
$G_{15}$ \cite{pontiki-etal-2015-semeval} &  & \cmark &  &  & \cmark &   &  \\
\hline
$SL_{15}$ \cite{pontiki-etal-2015-semeval}&  & \cmark &  & \cmark &  & \cmark   &  \\
\hline
$SR_{15}$ / $SH_{15}$ \cite{pontiki-etal-2015-semeval}&  & \cmark &  & \cmark & \cmark & \cmark   &  \\
\hline
$SR_{16}$ \cite{DBLP:conf/semeval/PontikiGPAMAAZQ16} &  & \cmark &  & \cmark & \cmark & \cmark   &  \\
\hline
$SR_{16}$ \cite{DBLP:conf/semeval/PontikiGPAMAAZQ16} &  & \cmark &  & \cmark &  & \cmark  &  \\
\hline
$F_{18}$ \cite{fiqa_2018} & \cmark &  &  & \cmark & \cmark &    &  \\
\hline
$M_{ate19}$  \cite{mams_2019}&  & \cmark &  &  & \cmark &    &  \\
\hline
$M_{acc19}$  \cite{mams_2019}& \cmark & \cmark &  & \cmark &  &   &  \\
\hline
$SS_{22}$ \cite{structured_sentiment_barns_2022}&  & \cmark & \cmark &  & \cmark &   & \cmark \\
\hline
$A_{i23}$ \cite{absia_2023} & \cmark & \cmark &  &  & \cmark &   &  \\
\hline
$ME_{23}$ \cite{memd_2023} &  & \cmark & \cmark & \cmark & \cmark & \cmark   &  \\
\hline 
\end{tabular}
}
\caption{Alignment of datasets with UOC as described by Eq.~\ref{eq: form_op}. It should be noted that none of the datasets have annotations corresponding to $q$ and $r$.}

\vspace{-10pt}
\label{tab:sem_data_alignment}
\end{table}
 
\subsubsection{Evaluation Dataset Creation}

\noindent We observe that none of the datasets have all the annotations required to address the UOCE task. The annotation of a training dataset for UOCE is a non-trivial task and is outside the scope of this task. To evaluate the UOCE methods, we extend the annotations of a sample of $ME_{23}$ dataset, creating a small evaluation dataset. $ME_{23}$ was selected based on its multi-domain characteristics and the substantial overlap of its pre-existing labels with opinion concepts, as illustrated in Table \ref{tab: concept_data_align}. The $ME_{23}$ dataset comprises five domains: Books, Clothing, Hotel, Restaurant and Laptop. 
The evaluation dataset comprises 20 randomly selected sub-samples from each domain, resulting in a combined benchmark of 100 data points. Subsequently, we extend the annotations to include \textbf{qualifier}, \textbf{reason}, \textbf{sentiment intensity} and \textbf{holder labels}. We finalized the extended annotations with a consensus between three expert annotators.
The characteristics of the evaluation dataset are depicted in the table \ref{tab:bench_characteristic}, including the number of modifications made to previously annotated labels ($\Delta$). The dataset will be released publicly on GitHub under the Apache 2.0 license.

\begin{table}[htbp]
\setlength\extrarowheight{2pt}
\addtolength{\tabcolsep}{.1pt}
\footnotesize
    \centering

    \begin{tabular}{r|c|c|c}
     \hline
     \textbf{Annotation} & \textbf{Total} &\textbf{Unique} & $\Delta$ \\
      \hline
        Sentences & 100 & 100 & 0 \\
        Opinions & 134 & 134 & 18\\\hline
        Sentiment Polarity ($sp$)& 134 & 3 &  10\\
        Sentiment Intensity ($si$)& 134 & 3 &  N/A\\
        Sentiment Expression ($se$)& 111 & 96 & 44\\\hline
        Target Entity ($te$)& 134 & 24 & 38 \\
        Aspect Category ($ac$)& 134 & 18 & 38 \\
        Aspect Term ($at$)& 102 & 73 & 42\\\hline
        Opinion Holder Span ($hs$)& 61 &  10 & N/A \\
        Opinion Holder Entity ($he$)& 134 & 3 & N/A\\\hline
        Qualifier ($q$) & 31 & 24 & N/A\\
        Reason ($r$)& 46 & 46 & N/A\\\hline
         
    \end{tabular}
    \caption{Benchmark Dataset Characteristics \\ $\Delta$ column represents the changes in existing annotations before extension.}
    \label{tab:bench_characteristic}
    \vspace{-7pt}
\end{table}

\subsection{Baseline Methods}
\label{sec:bline}
In the UOCE opinion tuple (see Eq.\ref{eq: form_op}) some of the opinion concepts are extracted spans ($at$, $se$, $hs$, $q$, $r$), some are discrete classes ($sp$, $si$) and the remaining ones are generative ($te$, $ac$, $he$).  LLMs are known to be competent at few-shot inference and have a task-agnostic architecture \cite{brown_et_al_2020}. Therefore, our baselines use LLMs to generatively predict all the opinion concepts (see Eq. \ref{eq: form_op}) in the input text. The following two prompt variations are used: 
\begin{enumerate}
    \item \textbf{Natural Language Prompt (NLPrompt)}: The natural language prompt comprises four distinct components: Definitions (D), which describes the opinion concepts; In-Context Examples (E), which provides examples of the input text with the expected output; Format guidelines (F), describes the expected layout of the generated output; and the Query, which contains the text input for opinion mining and a text cue to start generating. The content of the Query varies; however, its position at the end of the prompt remains fixed in all variations. We conduct the UOCE experiments with different D, E and F sequences using different LLMs.     
    \item \textbf{Ontology Prompt (OntoPrompt)}: The ontology prompt has a similar organisation to NLPrompt. The only difference is the use of an ontology serialisation format to describe the UOC instead of natural language (see Appendix \ref{appendix:owl_format_specification}). When conducting the experiments with OntoPrompt, we utilise various ontology languages to describe UOC in the prompt. 
\end{enumerate}
Once we extract the opinions generatively using LLMs, we report the component-level exact match f1 scores (\ref{sec:eval_metrics}). 

\subsection{Experimental Settings}
The experiments were conducted on a machine with two NVIDIA RTX A6000 48GB GPUs. We employ the following open-weight LLMs for the experiments: Gemma-2 (9B, 27B) \cite{llm_gemma} , Mistral 7B \cite{llm_mistral_7b}, Mixtral 8x7B \cite{llm_mixtral} and Llama-3.1 (8B, 70B) \cite{llm_llama}. Additionally, we use OpenAI's GPT-4o and GPT-4o-mini \cite{llm_gpt4} accessed through an API interface. For the open-weight LLMs, 4-bit quantization is used to enable GPU inference. The generation parameters were kept constant across all models. We use a temperature value of 0.0 to ensure the most deterministic generation; the number of new tokens generated was restricted to 512. All relevant code and results will be provided on GitHub to ensure reproducibility.

\section{Results and Discussion}
\vspace{-10pt}
\begin{table}[!ht]
    \setlength{\tabcolsep}{2.2pt}
    \centering
    \resizebox{0.5\textwidth}{!}{
    \begingroup

    \begin{tabular}{c@{\hspace{1pt}}|@{\hspace{1.8pt}}ccccccc}
    \hline
    \multirow{2}{*}{\textbf{Model}}  & \multicolumn{7}{c}{\textbf{F1 Scores}}\\
     & \textbf{DEF} & \textbf{DFE} & \textbf{EDF} & \textbf{EFD} & \textbf{FDE} & \textbf{FED}  & \textbf{$\mu \pm \sigma$}\\
    \hline
    Gemma2 27B & 57.7 & 55.92 & 56.77 & 56.77 & 55.15 & 53.64 & 55.99 $ \pm $ 1.44\\
    Gemma2 9B & 57.2 & 55.85 & 58.56 & 58.4 & 55.35 & 54.46 & 56.64 $ \pm $ 1.68\\
    GPT-4o & 58.46 & 55.58 & 59.12 & 59.33 & 57.55 & 56.76 & \textbf{57.8} $ \pm $ 1.46\\
    GPT-4o-Mini & 54.67 & 53.88 & 55.59 & 57.0 & 53.29 & 56.26 & 55.12 $ \pm $ 1.42\\
    Llama 3.1 70B & 46.9 & 46.02 & 48.04 & 44.14 & 44.86 & 46.27 & 46.04 $ \pm $ 1.4\\
    Llama 3.1 8B & 46.36 & 49.88 & 43.84 & 44.73 & 48.79 & 35.54 & 44.86 $ \pm $ 5.11\\
    Mistral 7B & 48.0 & 48.52 & 49.09 & 48.46 & 49.61 & 50.3 & 49.0 $ \pm $ 0.85\\
    Mixtral 8x7B & 49.63 & 50.57 & 51.84 & 51.26 & 49.6 & 50.98 & 50.65 $ \pm $ 0.9\\
    \hline
    \textbf{$\mu $} & 52.36 & 52.03 & \textbf{52.86} & 52.51 & 51.78 & 50.53 & \\
    \textbf{$ \pm \sigma$} & 5.17 & 3.8 & 5.53 & 6.19 & 4.24 & 6.97 & \\
    \hline 
    \end{tabular}
    
    \endgroup
    }

    \resizebox{0.5\textwidth}{!}{
    \begingroup
    \begin{tabular}{c@{\hspace{1pt}}|@{\hspace{1.55pt}}cccccccc}
    \hline
    \multirow{2}{*}{\textbf{Model}}  & \multicolumn{8}{c}{\textbf{F1 Scores}}\\
     & \textbf{jsonld} & \textbf{man} & \textbf{obo} & \textbf{owf} & \textbf{owx} & \textbf{rdfx} & \textbf{ttl} & \textbf{$\mu \pm \sigma$}\\
    \hline

    Gemma2 27B & 57.36 & 56.54 & 57.59 & 55.49 & 57.96 & 55.35 & 58.76 & 57.01 $ \pm $ 1.27\\
    Gemma2 9B & 54.66 & 54.75 & 54.12 & 43.68 & 54.18 & 44.48 & 54.77 & 51.52 $ \pm $ 5.09\\
    GPT-4o & 57.71 & 56.41 & 57.47 & 57.65 & 56.0 & 57.45 & 58.13 & \textbf{57.26} $ \pm $ 0.76\\
    GPT-4o-Mini & 55.26 & 54.38 & 52.71 & 53.94 & 54.31 & 53.72 & 53.74 & 54.01 $ \pm $ 0.78\\
    Llama 70B & 51.39 & 50.32 & 52.2 & 51.66 & 49.41 & 51.26 & 50.91 & 51.02 $ \pm $ 0.92\\
    Llama 8B & 49.59 & 50.91 & 49.39 & 49.04 & 49.42 & 50.38 & 49.31 & 49.72 $ \pm $ 0.67\\
    Mistral 7B & 49.07 & 47.97 & 47.91 & 47.45 & 48.52 & 47.25 & 47.27 & 47.92 $ \pm $ 0.68\\
    Mixtral 8x7B & 51.75 & 50.79 & 50.38 & 50.26 & 50.63 & 49.18 & 51.36 & 50.62 $ \pm $ 0.83\\
    \hline
    \textbf{$\mu $} & \textbf{53.35} & 52.76 & 52.72 & 51.15 & 52.55 & 51.13 & 53.03 & \\
    \textbf{$ \pm \sigma$} & 3.37 & 3.17 & 3.55 & 4.53 & 3.52 & 4.28 & 4.08 & \\
    \hline
    
    \end{tabular}
    \endgroup
    }
    
    \caption{Effect of Definition (D), Examples (E) and Format (F) Variations in NLPrompt (Top) and Effect of Different Ontology representation format for Concept Description (D) in Prompts (Bottom)}
    \label{tab:merged_ablations}
\end{table}
\vspace{-5pt}
The baselines for UOCE are obtained generatively with LLMs using NLPrompts and OntoPrompts. The F1-scores for different variations of NLPrompts are reported in the table \ref{tab:merged_ablations} (top). The E-D-F sequence exhibits the highest average F1 score (52.86) across all E, D, and F sequences.

Similarly, for OntoPrompt, the experiments were conducted with different variations of the OntoPrompt. The ontology serialisation formats that vary across this prompt are specified and briefly described in appendix \ref{appendix:owl_format_specification}. The F1 scores from these experiments are reported in appendix \ref{tab:merged_ablations} (bottom). We obtained the highest average F1 score for OntoPrompt using JSON-LD (i.e. JSON for Linked Data) to describe the UOC ontology in the prompt. We also conclude the best prompt-LLM combination with these results by looking at the mean values. For NLPrompt, the (E-D-F) variant performs the best, and GPT-4o performs the best overall. Similarly, for OntoPrompt, JSON-LD is the best-performing ontology serialisation format, and GPT-4o is the best-performing model. 

\subsection{Comparison with existing methods}
\begin{table}[h]
    \footnotesize
    \centering
    \vspace{-6pt}
    \begin{tabular}{r|l|c|c|c}
    
    \hline
    \multirow{2}{3em}{\textbf{Task}} &\multirow{2}{3em}{\textbf{Model}} &  \multicolumn{3}{c}{Component-Level EM}\\ \cline{3-5}
  
      & &  \textbf{P} &\textbf{R} & \textbf{F1} \\\hline
       \multirow{4}{3em}{ASTE}& GEN-SCL-NAT & 60.25 & 70.14 & 64.82\\
        & MVP  & 61.26 & 67.66 & 64.30\\
        & Ours (NLPrompt) & 75.24 & \textbf{74.15} & 74.69 \\
        & Ours (OntoPrompt) & \textbf{75.87} & 73.67 & \textbf{74.75}\\
       \hline
       \multirow{4}{3em}{ACOS} & GEN-SCL-NAT &  49.61 & 57.76 & 53.38\\
        & MVP & 52.83 & \textbf{58.35} & 55.46\\
        & Ours (NLPrompt) & 58.23 & 57.39 & \textbf{57.81} \\
        & Ours (OntoPrompt) & \textbf{58.35} & 56.67 & 57.49 \\
       \hline
        \multirow{4}{3em}{UOCE}& GEN-SCL-NAT & 39.10 & 45.52 & 42.07\\
        & MVP  & 35.60 & 39.32 & 37.37\\
        & Ours (NLPrompt) & \textbf{55.22} & \textbf{63.62} & \textbf{59.12} \\
        & Ours (OntoPrompt) & 53.9 & 62.1 & 57.71\\

    \hline         
    \end{tabular}
    \caption{Comparing baseline results with Component-Level Exact Match}
    \label{tab:uoce_comp}
    \vspace{-6pt}
\end{table}

 We compare the baseline methods with state-of-the-art (SOTA) ACOS and ASTE methods, as they are the most fine-grained forms of opinion extraction in the literature. ACOS contains 5 out of our 10 UOC labels and ASTE 3 out of 10 UOC labels. UOC concepts can be mapped to these tasks for comparison as: (i) ACOS corresponds to $o_{part}=(te,ac, at, ap,se)$, and (ii) ASTE to $o_{part}=(at, ap,se)$. 

The first SOTA model we consider is \textbf{GEN-SCL-NAT} \cite{gen_scl_nat_2022}, which improved the performance of generative ACOS models by addressing the limitations in identifying opinions with implicit sentiments. \textbf{Multi-View Prompting (MVP)} \cite{mvp_2023} improves on GEN-SCL-NAT by incorporating all the sub-ACOS tasks within a unified framework. It creates multiple training instances by manipulating the sequence of ACOS elements. 

Despite having a relatively lower F1 score (<60\%) for the UOCE task, we observe that the baseline methods outperform the state-of-the-art ASTE and ACOS tasks. The comparison results (Table \ref{tab:comp1}) illustrate the challenges UOCE poses and the benefits to other opinion mining formulations.

\subsection{Quantitative Analysis}
\paragraph{Overall Results}: In our UOCE experiments, GPT-4o had the highest F1 score of 59.33\% with an NLPrompt, closely followed by GPT-4o again with a prompt variation having an F1 score of 59.12\% also with an NLPrompt. OntoPrompt has the highest F1 score of 58.76\%, with Gemma-2 (27B), the third-highest overall score. 
 \paragraph{Effect of LLM Size}: For the same model, the version with a larger size performs better quantitatively on the evaluation dataset using the NLPrompt. However, we see some exceptions with the OntoPrompt. 
\paragraph{NLPrompt Vs OntoPrompt}: Although NLPrompt achieved the highest individual score, OntoPrompt demonstrated superior average values for the F1 score. Additionally, the results produced by OntoPrompt exhibited a lower standard deviation $\sigma$ of F1  scores, hinting at the higher robustness of OntoPrompt’s predictions.
\begin{table*}[h!]
  \centering
  \renewcommand{\tabcolsep}{1.3pt}
  \renewcommand{\arraystretch}{1.25}
  \footnotesize
  \begin{tabular}{r|c|c|c|c|l}
    \hline 
    \textbf{Extracted Labels} & \textbf{Ours (NLPrompt)} & \multicolumn{1}{c|}{\textbf{Ours (OntoPrompt)}} & \textbf{GEN-SCL-NAT} & \textbf{MVP} & \textbf{Gold Labels} \\
    \cline{2-3}
    \hline
    Aspect Term & \textcolor{red}{locations} & \textcolor{red}{location} & \textcolor{teal}{N/A} & \textcolor{teal}{N/A} & \textcolor{teal}{N/A} \\
    Aspect Category & \textcolor{teal}{general} & \textcolor{teal}{general} & \textcolor{teal}{general} & \textcolor{teal}{general} & \textcolor{teal}{general} \\
    Target Entity & \textcolor{red}{place} & \textcolor{teal}{location} & \textcolor{teal}{location} & \textcolor{red}{restaurant} & \textcolor{teal}{location} \\
    \hline
    Sentiment Expression & \textcolor{teal}{one of the best} & \textcolor{teal}{one of the best} & \textcolor{red}{best} & \textcolor{red}{best} & \textcolor{teal}{one of the best} \\
    Sentiment Polarity & \textcolor{teal}{positive} & \textcolor{teal}{positive} & \textcolor{teal}{positive} & \textcolor{teal}{positive} & \textcolor{teal}{positive} \\
    Sentiment Intensity & \textcolor{teal}{strong} & \textcolor{teal}{strong} & \xmark & \xmark & \textcolor{teal}{strong} \\
    \hline
    Holder Span  & \textcolor{teal}{N/A} & \textcolor{teal}{N/A} & \xmark & \xmark & \textcolor{teal}{N/A} \\
    Holder Entity & \textcolor{teal}{author} & \textcolor{teal}{author} & \xmark & \xmark & \textcolor{teal}{author} \\
    \hline
    \multirow{2}{*}{Qualifier} & \textcolor{red}{you could} & \multirow{2}{*}{\textcolor{red}{N/A}}& \multirow{2}{*}{\xmark}& \multirow{2}{*}{\xmark} & \multirow{2}{*}{\textcolor{teal}{stay at in Boston}}\\ 
    & \textcolor{teal}{stay at in Boston} &  & & & \\
    \hline
    Reason & \textcolor{teal}{N/A} & \textcolor{teal}{N/A} &\xmark &\xmark &\textcolor{teal}{N/A} \\
    \hline
  \end{tabular}
  \caption{Automatic Opinion Extraction for ``By far one of the best locations you could stay at in Boston.''}
  \label{tab:opinion_qual}
  \vspace{-10pt}
\end{table*}
\subsection{Qualitative Analysis}
In table \ref{tab:opinion_qual}, we discuss examples of UOCE outputs of different models for the sentence: \textit{By far one of the best locations you could stay at in Boston ."}. We see a high agreement of various opinion concepts extracted across the models. None of the models recognized the qualifier span it correctly. The error in falsely recognizing the aspect term highlights a lack of nuanced understanding of the aspect term when using in-context generative baselines. The GEN\_SCL\_NAT and MVP models were trained on ABSA datasets and do not have difficulty identifying aspect terms. Being trained on ACOS tasks, the GEN\_SCL\_NAT and MVP models cannot extract all the UOCE concepts.  LLMs struggle to recognize qualifiers and reasons in our benchmark dataset as they require nuanced semantic understanding. We believe there is ample room for improvement on the baselines by exploring methods of better semantic utilization.

\section{Conclusion}
This paper introduced the Unified Opinion Concepts (UOC) ontology, which integrates the diverse perspectives on opinion mining task descriptions in NLP based on \citet{DBLP:books/sp/mining2012/LiuZ12} and the ontological opinion representation \cite{marl_opinion_2011}. UOC formalizes the semantic structure of opinions previously expressed implicitly and scattered across the opinion-mining literature. We proposed Unified Opinion Concept Extraction (UOCE) as an NLP task based on the expressive semantics of the UOC ontology. To facilitate system development for UOCE, an evaluation dataset that extends the annotations of a gold standard dataset is also provided.  

We also introduced tailored evaluation metrics for the extracted opinions, comparing them with traditional metrics for fine-grained opinion-mining tasks. Finally, we provided baseline methods for UOCE using LLMs. We compared our baselines against comparable state-of-the-art methods approaches to the existing fine-grained opinion-mining task in the literature to highlight the complexity of UOCE. The comparison in Table \ref{tab:uoce_comp} indicates UOC formulation's potential benefits for other fine-grained opinion-mining tasks.

\section{Limitations and Future Work}
The Unified Opinion Concepts (UOC) ontology offers an expressive framework for semantically structured opinion mining, yet several limitations must be acknowledged. Firstly, the evaluation dataset provided is helpful for evaluation purposes but is insufficient in size to train a practical system using data-driven approaches. The only training data points we used for our baseline approaches were the in-context examples in the prompt. 

Secondly, even after incorporating element-wise exact matches, current evaluation metrics rely on overlapping extracted or generated opinion concepts with the gold labels. They penalize any lack of exact matching between predicted tokens and reference labels. This strictness mainly affects the evaluation of reasons and qualifiers, which often have considerable token spans. Therefore, adopting flexible and context-aware evaluation metrics would significantly benefit this research.

Lastly, the established baselines open significant scope for exploring effective machine learning techniques to enhance performance. Evaluating different modelling approaches, such as transfer learning and graph machine learning, is essential to understand better and utilize the comprehensive semantic structure introduced in this work.

\section*{Acknowledgments}
This work was conducted with the financial support of the Science Foundation Ireland (SFI) under Grant Number SFI/12/RC/2289\_P2 (Insight\_2) and was also supported by funding from the Irish Research Council (IRC) for the Postdoctoral Fellowship award GOIPD/2023/1556.

\appendix
\section{Effects of Metric Selection}
\label{sec:appendix_metric}

\begin{table}[!hbt]
\setlength\extrarowheight{2pt}
    \resizebox{0.5\textwidth}{!}{
    \begin{tabular}{r?l?l|l|l?l|l|l}
    \hline
    \multirow{2}{3em}{\textbf{TASK}} &\multirow{2}{3em}{\textbf{MODEL}} & \multicolumn{3}{l?}{\textbf{TUP-LEV EM}} & \multicolumn{3}{l}{\textbf{COM-LEV EM}}\\\cline{3-8}
  
      & & \textbf{P} &\textbf{R} & \textbf{F1} & \textbf{P} &\textbf{R} & \textbf{F1} \\
      \hline
       \multirow{4}{3em}{ASTE}& GEN-SCL-NAT& 32.68 & 37.31 & 34.84 & 60.25 & 70.14 & 64.82\\
        & MVP & 33.10 & 36.57 & 34.75 & 61.26 & 67.66 & 64.30\\
        & Ours (NLPrompt) & 38.75 & 44.92 & 41.61 & 75.24 & 74.15 & 74.69 \\
         & Ours (OntoPrompt) & 39.62 & 45.65 & 42.42 & 75.87 & 73.67 & 74.75\\
         \hline
         \multirow{4}{3em}{ACOS} & GEN-SCL-NAT& 3.20 & 3.73 & 3.45 & 49.61 & 57.76 & 53.38\\
         & MVP & 12.84 & 14.18 & 13.48 & 52.83 & 58.35 & 55.46\\
         & Ours (NLPrompt) & 4.37 & 5.07 & 4.89 & 58.23 & 57.39 & 57.81 \\
         & Ours (OntoPrompt) & 3.77 & 4.34 & 4.04 & 58.35 & 56.67 & 57.49 \\
         \hline
         \multirow{4}{3em}{UOCE}& GEN-SCL-NAT & \textcolor{red}{0.00} & \textcolor{red}{0.00} & \textcolor{red}{0.00} & 39.10 & 45.52 & 42.07\\
        & MVP  & \textcolor{red}{0.00} & \textcolor{red}{0.00} & \textcolor{red}{0.00} & 35.60 & 39.32 & 37.37\\
        & Ours (NLPrompt) & \textcolor{red}{0.00} & \textcolor{red}{0.00} & \textcolor{red}{0.00} &  \textbf{55.22} & \textbf{63.62} & \textbf{59.12} \\
        & Ours (OntoPrompt) & \textcolor{red}{0.00} & \textcolor{red}{0.00} & \textcolor{red}{0.00} & 53.9 & 62.1 & 57.71\\

    \hline         
    \end{tabular}
    }
    \caption{Comparing baseline results using Tuple-level Exact Match (\textbf{TUP-LEV EM}) and Component-Level Exact Match (\textbf{COM-LEV EM})}
    \label{tab:comp1}
    \vspace{-8pt}
\end{table}

\noindent As evident from Table \ref{tab:comp1}, due to the stringency of the Tuple-Level Exact Match metric used by opinion mining systems, it fails to measure the output of the extraction systems capable of partial opinion extraction.

This discontinuity in measurement becomes even more apparent as the multi-extraction tasks get more challenging from ASTE to ACOS until it eventually fails to measure anything for the UOCE task (i.e. no of elements to be extracted increases).

\section{Ontology Serialisation Documentation Links}
\label{appendix:owl_format_specification}
The links to the documentation of various ontology serialization formats used are listed below:

\begin{table}[ht]
\setlength\extrarowheight{1.2pt}
\scriptsize
\begin{tabularx}{\linewidth}{l X}
\textbf{Syntax/Format} & \textbf{Reference URL} \\
\hline
JSON-LD & \url{https://json-ld.org/} \\
Manchester OWL Syntax & \url{https://www.w3.org/TR/owl2-manchester-syntax/} \\
OWL/XML Syntax & \url{https://www.w3.org/TR/owl-xmlsyntax/} \\
OBO Format & \url{https://owlcollab.github.io/oboformat/doc/GO.format.obo-1_4.html} \\
OWL Functional Syntax & \url{https://www.w3.org/TR/owl2-syntax/} \\
RDF/XML Syntax & \url{https://www.w3.org/TR/rdf-syntax-grammar/} \\
Turtle Syntax & \url{https://www.w3.org/TR/turtle/} \\
\end{tabularx}
\end{table}

\end{document}